\documentclass{article}

\usepackage[final]{nips_2018}

\usepackage[utf8]{inputenc} 
\usepackage[T1]{fontenc}    
\usepackage{hyperref}       
\usepackage{amsfonts}       
\usepackage{nicefrac}       
\usepackage{microtype}      
\usepackage{hyperref}
\usepackage{xcolor}
\hypersetup{colorlinks, citecolor={blue!50!black}, urlcolor={blue!50!black}}
\usepackage{todonotes}
\usepackage{tikz}
\usetikzlibrary{bayesnet}
\usepackage{amsfonts,amsmath,amssymb}
\usepackage{mathtools}
\usepackage{xparse}
\usepackage{dsfont}

\let\originalleft\left
\let\originalright\right
\renewcommand{\left}{\mathopen{}\mathclose\bgroup\originalleft}
\renewcommand{\right}{\aftergroup\egroup\originalright}

\newcommand{\cond}{\mathbin{|}}
\newcommand{\midcond}{\,\middle|\,}

\NewDocumentCommand\expec{somg}{
	{\mathbb{E}%
	\IfValueT{#2}%
		{_{#2}}%
	\IfBooleanTF{#1}
		{%
			\!\left[#3%
			\IfValueT{#4}%
				{\midcond #4}%
			\right]
		}{%
			[#3%
			\IfValueT{#4}%
				{\cond #4}%
			]%
		}%
	}%
}

\NewDocumentCommand\variance{somg}{
	{\mathbb{V}%
	\IfValueT{#2}%
		{_{#2}}%
	\IfBooleanTF{#1}
		{%
			\!\left[#3%
			\IfValueT{#4}%
				{\midcond #4}%
			\right]
		}{%
			[#3%
			\IfValueT{#4}%
				{\cond #4}%
			]%
		}%
	}%
}

\NewDocumentCommand\kldiv{smm}{
	{\mathit{KL}%
	\IfBooleanTF{#1}
		{%
			\left[#2%
			\IfValueT{#3}%
				{\middle\Vert #3}%
			\right]
		}{%
			[#2%
			\IfValueT{#3}%
				{\Vert #3}%
			]%
		}%
	}%
}

\renewcommand\*[1]{%
	\ifcat\noexpand#1\relax 
   		\boldsymbol{#1}
	\else
		\mathbf{#1}
	\fi
}

\tikzset{const/.append style={circle,minimum size=10pt}}

\NewDocumentCommand{\CondBracketsNoStar}{mg}{%
	(#1\IfValueT{#2}{\cond #2})%
}
\NewDocumentCommand{\CondBracketsStar}{mg}{%
	\left(#1\IfValueT{#2}{\midcond #2}\right)%
}
\NewDocumentCommand{\CondBrackets}{smg}{%
	\IfBooleanTF{#1}{\CondBracketsStar{#2}{#3}}{\CondBracketsNoStar{#2}{#3}}%
}

\newcommand{\prob}{p\CondBrackets}

\usepackage[capitalise]{cleveref} 

\makeatletter
\if@nipsfinal
    \renewcommand{\@noticestring}{NeurIPS \@nipsyear\/ Workshop on All of Bayesian Nonparametrics (BNP$@$NeurIPS \@nipsyear), \@nipslocation.}
\fi
\makeatother

\newcommand{\DP}{\operatorname{DP}}
\newcommand{\GEM}{\operatorname{GEM}}
\newcommand{\HGEM}{\operatorname{HGEM}}

\newcommand{\Cat}{\operatorname{Cat}}
\newcommand{\Dir}{\operatorname{Dir}}

\newcommand{\ObsPrior}{H_\mathrm{X}}
\newcommand{\ObsLik}{F_\mathrm{X}}

\newcommand{\LabelPrior}{H_\mathrm{Y}}
\newcommand{\LabelLik}{F_\mathrm{Y}}

\newcommand{\LabelPost}{\widehat{\LabelPrior}}
\newcommand{\LabelBase}{L}

\newcommand{\EnvPrior}{H_\mathrm{E}}
\newcommand{\EnvLik}{F_\mathrm{E}}

\newcommand{\LabelledIndices}{\mathcal{L}}

\def\eg{e.g.~}
\def\ie{i.e.~}

\title{Contextual Face Recognition with a Nested-Hierarchical Nonparametric Identity Model}

\author{
    Daniel C.~Castro\thanks{Work partly done during an internship at Microsoft Research.} \\
    Imperial College London, UK \\
    \texttt{dc315@imperial.ac.uk} \\
\And
    Sebastian Nowozin \\
    Microsoft Research Cambridge, UK \\
    \texttt{Sebastian.Nowozin@microsoft.com}
}

\begin{document}

\maketitle

\begin{abstract}
Current face recognition systems typically operate via classification into known identities obtained from supervised identity annotations.
There are two problems with this paradigm:
(1) current systems are unable to benefit from often abundant unlabelled data; and
(2) they equate successful recognition with labelling a given input image.
Humans, on the other hand, regularly perform identification of individuals completely unsupervised, recognising the identity of someone they have seen before even without being able to name that individual.
How can we go beyond the current classification paradigm towards a more human understanding of identities?
%
In previous work, we proposed an integrated Bayesian model that coherently reasons about the observed images, identities, partial knowledge about names, and the situational context of each observation.
Here, we propose extensions of the contextual component of this model, enabling unsupervised discovery of an unbounded number of contexts for improved face recognition.
\end{abstract}

\section{Introduction}

Face identification can be decomposed into two sub-problems: \emph{recognition} and \emph{tagging}. Here we understand recognition as the unsupervised task of matching an observed face to a cluster of previously seen faces with similar appearance (disregarding variations in pose, illumination etc.), which we refer to as an \emph{identity}. Humans routinely operate at this level of abstraction to recognise familiar faces: even when people's names are not known, we can still tell them apart. Tagging, on the other hand, refers to putting names to faces, \ie associating string literals to known identities.

\begin{figure}
    \centering
    \begin{minipage}[c]{.4\linewidth}
        \centering
        \includegraphics[width=\linewidth]{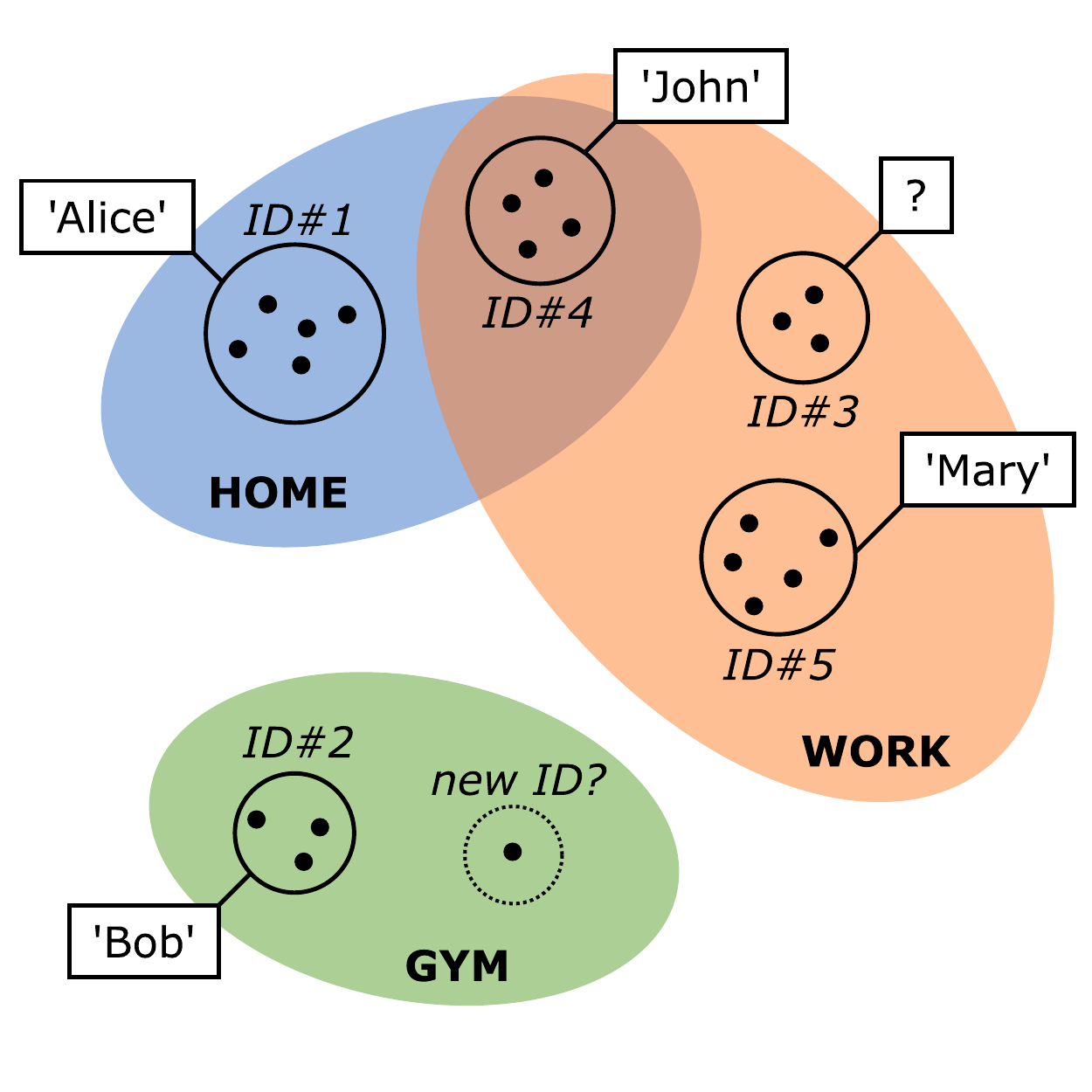}
        \caption{Context-aware model of identities \citep{Castro2018}}
        \label{fig:setting}
    \end{minipage}\hfill
    \begin{minipage}[c]{.55\linewidth}
        \centering
        \tikzset{>=latex}
\tikzstyle{wrap} += [inner sep=1pt]
\tikzstyle{plate} += [inner sep=4pt]
\tikzstyle{const} += [inner sep=1pt, node distance=.5]
\tikzset{
    diag fill/.style 2 args={fill=#2, path picture={
        \fill[#1, sharp corners] (path picture bounding box.south west) -|
        (path picture bounding box.north east) -- cycle;}}
}
\tikzstyle{partial} = [latent, diag fill={gray!25}{white}]
\tikzstyle{marg} = [latent, dashed]

\begin{tikzpicture}[x=36pt, y=48pt]
	\node[obs]		at ( 0, 0)	(x)		{$\*x_n$};
	\node[latent]	at (-1, 1)	(z)		{$z_n$};
	\node[latent]	at (-2, 0)	(c)		{$c_m$};
	\node[partial]	at ( 0, 1)	(y)		{$y_n$};
	\node[latent]	at (-2,-1)	(omega)	{$\*\omega$};
	\node[latent]	at (-2, 1)	(pi)	{$\*\pi^*_c$};
	\node[latent]	at (-1.7,1.8) (chi)	{$\*\pi_0$};
	\node[latent]	at ( 1, 0)	(theta)	{$\theta^*_i$};
	\node[latent]	at ( 1, 1)	(l)		{$y^*_i$};
	\node[const]	at (-.9,1.8)(alpha) {$\alpha_0$};
	\node[const]	at (-2.3,1.8)(alphac){$\alpha$};
	\node[const]	at (-2.8,-1)(gamma)	{$\gamma_0$};
	\node[const]	at ( 2, 0)	(hx)	{$\ObsPrior$};
	\node[latent]	at ( 2, 1)	(hy)	{$\LabelPrior$};
    \node[const]	at (-1, 0)	(f)		{$f_n$};
    \node[obs]      at (-3, 0)  (e)     {$\*e_m$};
    \node[latent]   at (-3, 1)  (eta)   {$\eta^*_c$};
    \node[const]    at (-3,1.8) (he)    {$\EnvPrior$};

	\edge {z, theta}	{x};
	\edge {c, pi, f}	{z};
	\edge {omega}		{c};
	\edge {gamma}		{omega};
	\edge {alpha}		{chi};
	\edge {alphac, chi}	{pi};
	\edge {hx}			{theta};
	\edge {z, l}	    {y};
	\edge {hy}			{l};
	\edge {c, eta}      {e};
	\edge {he}          {eta};
	
    \plate {N} {(z)(x)(y)(f)}   {$N$};
    \plate {M} {(c)(e)}		    {$M$};
	\plate {C} {(pi)(eta)}      {$\vphantom{N}\infty$};
	\plate {I} {(theta)(l)}     {$\vphantom{N}\infty$};
    
    \node[left=2pt of C,rotate=90,anchor=south] {context $c$};
    \node[left=2pt of M,rotate=90,anchor=south] {frame $m$};
    \node[below=2pt of N] {observation $n$};
    \node[below=2pt of I] {identity $i$};
\end{tikzpicture}
        \caption{Overview of the proposed probabilistic model}
        \label{fig:diagram}
    \end{minipage}
\end{figure}

An important aspect of social interactions is that, as an individual continues to observe faces every day, they encounter some people much more often than others, and the total number of distinct identities ever met tends to increase virtually without bounds. Additionally, we argue that human face recognition does not happen in an isolated environment, but situational contexts (\eg `home', `work', `gym') constitute strong cues for the groups of people a person expects to meet (\cref{fig:setting}).

With regards to tagging, in daily life we very rarely obtain named face observations: acquaintances normally introduce themselves only once, and not repeatedly whenever they are in our field of view. In other words, humans are naturally capable of semi-supervised learning, generalising sparse name annotations to all observations of the corresponding individuals, while additionally reconciling naming conflicts due to noise and uncertainty.

We recently introduced a unified Bayesian model which reflects all the above considerations on identity distributions, context-awareness and labelling (\cref{fig:setting}) \citep{Castro2018}. Our nonparametric identity model effectively represents an unbounded population of identities, while taking contextual co-occurrence relations and sparse noisy labels into account.

In this preliminary work, we extend the referred context model in two ways: we explore its limit with an unbounded number of contexts, uncovering a rich nonparametric structure, and we lay the foundations for incorporating environmental cues (such as timestamps and geographical locations of frames) in our model to improve unsupervised context discovery and prediction.

\section{Background}

We begin by reviewing the face identification framework presented in \citet{Castro2018}, consisting of four main components: a context model (which we extend in \cref{sec:new_context}), an identity model, a face model, and a semi-supervised label model.

\subsection{Context Model}\label{sec:old_context}
Data is assumed to be collected in frames, \ie photo album or video stills, which are run through some off-the-shelf face detector. This produces $N$ observations, grouped into the $M$ frames via an indicator $f_n=m$ for each observation $n$ in frame $m$. Context is therefore naturally shared among all face detections in each frame. We model context as a discrete latent variable, representing categories of situations in which a subject may find herself: \eg home, work, gym.

We assume the context indicators $c_m \in \{1,\dots,C\}$ for each frame $m$, where $C$ is some fixed number of distinct contexts, are independently distributed according to probabilities $\*\omega$, which themselves follow a Dirichlet prior:
\begin{align}
	\*\omega				&\sim \Dir(\*\gamma) \,, &
	c_m \cond \*\omega		&\sim \Cat(\*\omega) \,, \quad m = 1,\dots,M \,,
\end{align}
where $M$ is the total number of frames. In our simulation experiments in \citet{Castro2018}, we used a symmetric Dirichlet prior, setting $\*\gamma = (\gamma_0 / C, \dots, \gamma_0 / C)$.

\subsection{Identity Model}

In a daily-life scenario, an increasing number of unique identities will tend to appear as more faces are observed, \ie we do not expect a user to run out of new people to meet. Moreover, some people are likely to be encountered much more often than others. Since a Dirichlet process (DP) \citep{Ferguson1973} displays properties that mirror all of the above phenomena \citep{Teh2010}, it is a sound choice for modelling the distribution of identities.

Furthermore, the assumption that all people can potentially be encountered in any context, but with different probabilities, is perfectly captured by a hierarchical Dirichlet process (HDP) \citep{Teh2006}. Making use of the context model, we define one DP \emph{per context} $c$, each with concentration parameter $\alpha_c$ and sharing the same \emph{global} DP as a base measure. This hierarchical construction thus produces context-specific distributions over a common set of identities. Such a nonparametric model is additionally well suited for an open-set identification task, as it can elegantly estimate the prior probability of encountering an unknown identity.

To each of the $N$ face detections is associated a latent identity indicator variable, $z_n$. Letting $\*\pi_0$ denote the global identity distribution and $(\*\pi^*_c)_{c=1}^C$ the context-specific identity distributions, we can write the generative process as%
\begin{alignat}{2}
	\*\pi_0 			    &\sim \GEM(\alpha_0) \,, \\
	\*\pi^*_c \cond \*\pi_0 &\sim \HGEM(\alpha_c, \*\pi_0) \,, \quad	& c &= 1,\dots,C \,, \\
	z_n \cond f_n = m, \*c, (\*\pi^*_c)_c &\sim \Cat(\*\pi^*_{c_m})\,,  & n &= 1,\dots,N \,,
\end{alignat}
where $\GEM(\alpha_0)$ is the DP stick-breaking distribution, ${\pi_{0i} = \beta_{0i} \prod_{j=1}^{i-1} (1-\beta_{0j})}$, with ${\beta_{0i} \sim \operatorname{Beta}(1,\alpha_0)}$ and $i=1,\dots,\infty$ \citep{Sethuraman1994,Pitman2006}. We additionally define a \emph{hierarchical GEM distribution}, $\HGEM(\alpha, \*\pi_0)$, such that ${\pi^*_{ci} = \beta_{ci} \prod_{j=1}^{i-1} (1-\beta_{cj})}$, with ${\beta_{ci} \sim \operatorname{Beta}(\alpha_c \pi_{0i}, \alpha_c (1-\sum_{j=1}^i \pi_{0j}))}$ \citep[Eq.~(21)]{Teh2006}.

\subsection{Face Model}

We assume that the observed features of the $n$\textsuperscript{th} face, $\*x_n$, arise from a parametric family of distributions, $\ObsLik$. The parameters of this distribution, $\theta^*_i$, drawn from a prior, $\ObsPrior$, are unique for each identity and are shared across all face feature observations of the same person:%
\begin{align}
	\theta^*_i \sim \ObsPrior \,, \quad i = 1,\dots,\infty \,, &&
	\*x_n \cond z_n, \*\theta^* \sim \ObsLik(\theta^*_{z_n}) \,, \quad n = 1,\dots,N \,.
\end{align}
As a consequence, the marginal distribution of faces is given by an \emph{infinite mixture model} \citep{Antoniak1974}: $\prob{\*x_n}{c_n=c, \*\theta^*, \*\pi^*_c} = \sum_{i=1}^\infty \pi^*_{ci} \ObsLik(\*x_n \cond \theta^*_i)$.

In face recognition applications, it is typically more convenient and meaningful to extract a compact representation of face features than to work directly in a high-dimensional pixel space. For the experiments reported in \citet{Castro2018}, we used embeddings produced by a pre-trained neural network \citep{Amos2016}. We chose isotropic Gaussian mixture components for the face features ($\ObsLik$), with an empirical Gaussian--inverse gamma prior for their means and variances ($\ObsPrior$).

\subsection{Label Model}

We expect to work with only a small number of user-labelled observations. Building on the \emph{cluster assumption} for semi-supervised learning \cite[Sec.~1.2.2]{Chapelle2006}, we attach a label variable (a \emph{name}) to each cluster (identity), here denoted $y^*_i$. Since the number of distinct labels will tend to increase without bounds as more data is observed, we adopt a further nonparametric prior on these identity-wide labels, $\LabelPrior$,%
\footnote{One could instead consider a Pitman--Yor process if power-law behaviour seems more appropriate than the DP's exponential tails \citep{Pitman1997}.}
using some base probability distribution $\LabelBase$ over the countable but unbounded label space (\eg strings). In \citet{Castro2018} we defined $\LabelBase$ over a rudimentary language model. Lastly, the observed labels, $y_n$, are assumed potentially corrupted through some noise process, $\LabelLik$. Let $\LabelledIndices$ denote the set of indices of the labelled data. We then have
\begin{alignat}{2}
	\LabelPrior				&\sim \DP(\lambda, \LabelBase) \,, \\
	y^*_i \cond \LabelPrior	&\sim \LabelPrior \,,	& i &= 1,\dots,\infty \,, \\
	y_n \cond z_n, \*y^*, \LabelPrior	&\sim \LabelLik(y^*_{z_n}; \LabelPrior) \,,	\quad		& n &\in \LabelledIndices \,.
\end{alignat}

All concrete knowledge we have about the random label prior $\LabelPrior$ comes from the set of observed labels, $\*y_\LabelledIndices$. Crucially, we can easily marginalise out $\LabelPrior$ \citep{Teh2010}, obtaining a tractable predictive label distribution, $\LabelPost(y^*_{I+1} \cond \*y^*)$.

According to the proposed noise model, an observed label, $y_n$, agrees with its identity's assigned label, $y^*_{z_n}$, with a fixed probability. Otherwise, it is assumed to come from a modified label distribution, in which we delete $y^*_{z_n}$ from $\LabelPost$ and renormalise it. Here we use $\LabelPost$ in the error distribution instead of $\LabelBase$ to reflect that a user is likely to mistake a person's name for another known name, rather than for an arbitrary random string.

\section{Extended Context Model}\label{sec:new_context}

The context framework employed in \citet{Castro2018} assumes a finite collection of pre-specified contexts and is fully supervised: an explicit context label is observed with each frame. This simplified scenario was adopted as a proof of concept, yet is admittedly unrealistic.

\subsection{Unbounded Contexts}

As reviewed in \cref{sec:old_context}, the original context model had a \emph{finite} $\Dir(\tfrac{\gamma_0}{C}, \dots, \tfrac{\gamma_0}{C})$ prior. A natural extension of such model is to take its limit as $C\to\infty$, while tying the values of all context-wise concentration hyperparameters ($\alpha_c = \alpha, \forall c$), which results in a Dirichlet process \citep{Neal2000}. In particular, up to a reordering of the contexts, the prior on context proportions, $\*\omega$, becomes $\GEM(\gamma_0)$.

This transformation has interesting theoretical and practical implications: the resulting structure is a \emph{nested-hierarchical Dirichlet process}.%
\footnote{This is related to the dual-HDP described in \citet{Wang2009b} and the single-entity model of \citet{Agrawal2013}, for example, although these works tended to focus on textual topic modelling.}
As before, at the top level we have the global identity distribution, $G_0$, over face parameters and labels, and the context-specific identity distributions, $(G^*_c)_{c=1}^C$, follow a DP with $G_0$ as a base measure:
\begin{alignat}{2}
    G_0 \cond \LabelPrior   &\sim \DP(\alpha_0, \ObsPrior \otimes \LabelPrior) \,, \quad \\
    G^*_c \cond G_0         &\sim \DP(\alpha, G_0) \,, & c &= 1,\dots,\infty \,, \label{eq:HDP_bottom}
\end{alignat}
a prototypical example of a hierarchical DP (HDP) \citep{Teh2006}. If $G_0 = \sum_{i=1}^\infty \pi_{0i} \delta_{(\theta^*_i, y^*_i)}$, we can write $G^*_c = \sum_{i=1}^\infty \pi^*_{ci} \delta_{(\theta^*_i, y^*_i)}$.

Now, the nonparametric distribution of contexts implies wrapping the bottom level of the HDP, \cref{eq:HDP_bottom}, as base for another DP, to form a nested DP \citep{Blei2010,Rodriguez2008}:%
\begin{alignat}{2}
    Q \cond G_0 &\sim \DP(\gamma_0, \DP(\alpha, G_0)) \,, \label{eq:nDP} \quad \\
    G_m \cond Q &\sim Q = \textstyle \sum_{c=1}^\infty \omega_c \delta_{G^*_c} \,, & m &= 1,\dots,M \,.
\end{alignat}
This construction inherits desirable properties from both elements: the hierarchy ensures that all frame-wise identity distributions, $(G_m)_{m=1}^M$, have the same support, and nesting produces clusters of frames with shared identity weights (\ie contexts).

\subsection{Environmental Cues}

While a purely identity-driven unsupervised context model may be able to disentangle co-occurrence patterns given enough data, we believe that environmental cues---such as timestamp and GPS coordinates of an acquired frame, if available---could considerably facilitate context discovery and prediction, in turn improving inference about identities.

Let us define $\*e_m$ as the environmental measurements available for frame $m$, $\EnvLik$ a likelihood family parametrised by $\eta_m$, and $\EnvPrior$ a prior distribution for such parameters. Plugging $\DP(\alpha, G_0) \otimes \EnvPrior$ as base measure for the nested DP $Q$ in \cref{eq:nDP}, we can write
\begin{align}
    \eta^*_c \sim \EnvPrior \,, \quad c=1,\dots,\infty \,, &&
    \*e_m \cond c_m, \*\eta^* \sim \EnvLik(\eta^*_{c_m}) \,, \quad m=1,\dots,M \,.
\end{align} 
Some preliminary ideas for a spatial model include a `geodetic' Fisher distribution or a tangential Gaussian \citep{Straub2015}, while a temporal model would have to accommodate recurring and occasional contexts, potentially adopting a Cox process formalism \citep{Cox1955}. 

\section{Conclusion}

In this work, we reviewed the fully Bayesian treatment introduced in \citet{Castro2018} of the face identification problem. Each component of our proposed approach was motivated from human intuition about face recognition and tagging in daily social interactions, such that our principled identity model can contemplate context-specific probabilities of meeting an unbounded population.

We further proposed a nonparametric extension of the context model enabling unbounded context discovery, and discussed some of its theoretical implications in terms of nested-hierarchical nonparametric structures. Finally, we briefly examined how available environmental cues could be integrated into the model to replace the simplified supervised setting.

\subsubsection*{Acknowledgments}
This work was partly supported by CAPES, Brazil (BEX 1500/2015-05).

\small
\bibliographystyle{hapalike_modified}
\bibliography{references,related}

\begin{thebibliography}{}

\bibitem[Agrawal et~al., 2013]{Agrawal2013}
Agrawal, P., Tekumalla, L.~S., and Bhattacharya, I. (2013).
\newblock Nested hierarchical {D}irichlet process for nonparametric
  entity-topic analysis.
\newblock In {\em Machine Learning and Knowledge Discovery in Databases -- ECML
  PKDD 2013}, volume 8189 of {\em LNCS}, pages 564--579. Springer, Berlin,
  Heidelberg.

\bibitem[Amos et~al., 2016]{Amos2016}
Amos, B., Ludwiczuk, B., and Satyanarayanan, M. (2016).
\newblock {OpenFace}: A general-purpose face recognition library with mobile
  applications.
\newblock Technical Report CMU-CS-16-118, CMU School of Computer Science.

\bibitem[Antoniak, 1974]{Antoniak1974}
Antoniak, C.~E. (1974).
\newblock Mixtures of {D}irichlet processes with applications to {B}ayesian
  nonparametric problems.
\newblock {\em The Annals of Statistics}, 2(6):1152--1174.

\bibitem[Blei et~al., 2010]{Blei2010}
Blei, D.~M., Griffiths, T.~L., and Jordan, M.~I. (2010).
\newblock The nested {C}hinese restaurant process and {B}ayesian nonparametric
  inference of topic hierarchies.
\newblock {\em Journal of the ACM}, 57(2):7.

\bibitem[Castro and Nowozin, 2018]{Castro2018}
Castro, D.~C. and Nowozin, S. (2018).
\newblock From face recognition to models of identity: A {B}ayesian approach to
  learning about unknown identities from unsupervised data.
\newblock In {\em Computer Vision -- ECCV 2018}, volume 11206 of {\em LNCS},
  pages 745--761. Springer.
\newblock Extended version with supplement: arXiv:1807.07872.

\bibitem[Chapelle et~al., 2006]{Chapelle2006}
Chapelle, O., Sch{\"{o}}lkopf, B., and Zien, A., editors (2006).
\newblock {\em Semi-Supervised Learning}.
\newblock MIT Press.

\bibitem[Cox, 1955]{Cox1955}
Cox, D.~R. (1955).
\newblock Some statistical methods connected with series of events.
\newblock {\em Journal of the Royal Statistical Society: Series B
  (Methodological)}, 17(2):129--164.

\bibitem[Ferguson, 1973]{Ferguson1973}
Ferguson, T.~S. (1973).
\newblock A {B}ayesian analysis of some nonparametric problems.
\newblock {\em The Annals of Statistics}, 1(2):209--230.

\bibitem[Neal, 2000]{Neal2000}
Neal, R.~M. (2000).
\newblock {M}arkov chain sampling methods for {D}irichlet process mixture
  models.
\newblock {\em Journal of Computational and Graphical Statistics},
  9(2):249--265.

\bibitem[Pitman, 2006]{Pitman2006}
Pitman, J. (2006).
\newblock {\em Combinatorial Stochastic Processes}, volume 1875 of {\em Lecture
  Notes in Mathematics}.
\newblock Springer-Verlag, Berlin.
\newblock Lectures from the 32nd Summer School on Probability Theory held in
  Saint-Flour, July 7--24, 2002, with a foreword by Jean Picard.

\bibitem[Pitman and Yor, 1997]{Pitman1997}
Pitman, J. and Yor, M. (1997).
\newblock The two-parameter {P}oisson--{D}irichlet distribution derived from a
  stable subordinator.
\newblock {\em The Annals of Probability}, 25(2):855--900.

\bibitem[Rodr{\'{i}}guez et~al., 2008]{Rodriguez2008}
Rodr{\'{i}}guez, A., Dunson, D.~B., and Gelfand, A.~E. (2008).
\newblock The nested {D}irichlet process.
\newblock {\em Journal of the American Statistical Association},
  103(483):1131--1154.

\bibitem[Sethuraman, 1994]{Sethuraman1994}
Sethuraman, J. (1994).
\newblock A constructive definition of {D}irichlet priors.
\newblock {\em Statistica Sinica}, 4(2):639--650.

\bibitem[Straub et~al., 2015]{Straub2015}
Straub, J., Chang, J., Freifeld, O., and {Fisher III}, J.~W. (2015).
\newblock A {D}irichlet process mixture model for spherical data.
\newblock In {\em Proceedings of the 18th International Conference on
  Artificial Intelligence and Statistics (AISTATS 2015)}, volume~38 of {\em
  PMLR}, pages 930--938. PMLR.

\bibitem[Teh, 2010]{Teh2010}
Teh, Y.~W. (2010).
\newblock {D}irichlet process.
\newblock In Sammut, C. and Webb, G.~I., editors, {\em Encyclopedia of Machine
  Learning}, pages 280--287. Springer US.

\bibitem[Teh et~al., 2006]{Teh2006}
Teh, Y.~W., Jordan, M.~I., Beal, M.~J., and Blei, D.~M. (2006).
\newblock Hierarchical {D}irichlet processes.
\newblock {\em Journal of the American Statistical Association},
  101(476):1566--1581.

\bibitem[Wang et~al., 2009]{Wang2009b}
Wang, X., Ma, X., and Grimson, W. E.~L. (2009).
\newblock Unsupervised activity perception in crowded and complicated scenes
  using hierarchical {B}ayesian models.
\newblock {\em IEEE Transactions on Pattern Analysis and Machine Intelligence},
  31(3):539--555.

\end{thebibliography}

\end{document}